%% file: main.tex
\documentclass[10pt,twocolumn,letterpaper]{article}

\usepackage{iccv}
\usepackage{times}
\usepackage{epsfig}
\usepackage{graphicx}
\usepackage{amsmath}
\usepackage{amssymb}

\usepackage{booktabs}

\usepackage{bm}
\usepackage{algorithm}
\usepackage{multirow}
\usepackage{wrapfig}
\usepackage{amsthm}
\usepackage{caption}
\usepackage{listings}
\usepackage{algpseudocode}
\usepackage[T1]{fontenc}
\usepackage{xcolor}
\usepackage{colortbl}

\newcolumntype{x}[1]{>{\centering\arraybackslash}p{#1pt}}
\newcolumntype{y}[1]{>{\raggedright\arraybackslash}p{#1pt}}
\newcolumntype{z}[1]{>{\raggedleft\arraybackslash}p{#1pt}}
\definecolor{Highlight}{HTML}{39b54a}  
\newcommand{\hl}[1]{\textcolor{Highlight}{\textbf{#1}}}
\newcommand{\dtplus}[2]{\textbf{#1}\fontsize{6.5pt}{0.1em}\selectfont~\!\hl{(+#2)}}

\usepackage{array}
\renewcommand{\paragraph}[1]{\vspace{1.25mm}\noindent\textbf{#1}}
\newcommand{\tablestyle}[2]{\setlength{\tabcolsep}{#1}\renewcommand{\arraystretch}{#2}\centering\footnotesize}
\usepackage{pifont}

\newcommand{\OurMethod}{Deep Incubation}
\newlength\savewidth\newcommand\shline{\noalign{\global\savewidth\arrayrulewidth
\global\arrayrulewidth 1pt}\hline\noalign{\global\arrayrulewidth\savewidth}}
\usepackage[citecolor=citecolor,pagebackref=true,breaklinks=true,letterpaper=true,colorlinks,bookmarks=false]{hyperref}

 \iccvfinalcopy 

\def\iccvPaperID{4671} 

\ificcvfinal\fi

\definecolor{citecolor}{HTML}{0071BC}
\definecolor{grey_tbl}{HTML}{e6e6e6}
\definecolor{linkcolor}{HTML}{ED1C24}
\definecolor{up}{HTML}{39b54a}

\usepackage[capitalize]{cleveref}
\crefname{section}{Sec.}{Secs.}
\Crefname{section}{Section}{Sections}
\Crefname{table}{Table}{Tables}
\crefname{table}{Tab.}{Tabs.}

\begin{document}

\title{Deep Incubation: Training Large Models by Divide-and-Conquering}

\author{
Zanlin Ni$^{1}$\thanks{Equal contribution.}\ \ \ \
Yulin Wang$^{1*}$ \ \ \
Jiangwei Yu$^{1}$ \ \ \
Haojun Jiang$^{1}$\ \ \
Yue Cao$^{2}$ \ \ \
Gao Huang$^{1, 2}$\thanks{Corresponding author.} \\
$^{1}$Department of Automation, BNRist, Tsinghua University, Beijing, China \\
$^{2}$Beijing Academy of Artificial Intelligence, Beijing, China\\
\texttt{\footnotesize \{nzl22, wang-yl19, yu-jw19, jhj20\}@mails.tsinghua.edu.cn,} \\
\texttt{\footnotesize caoyue10@gmail.com,} \texttt{\footnotesize gaohuang@tsinghua.edu.cn}
}
\maketitle

\input{abstract.tex}
\input{intro.tex}
\input{related.tex}
\input{method.tex}
\input{experiments.tex}
\input{conclusion.tex}
\input{acknowledgement.tex}
{\small
\bibliographystyle{ieee_fullname}
\bibliography{main}
}

\end{document}

%% file: abstract.tex
\begin{abstract}
Recent years have witnessed a remarkable success of large deep learning models.
However, training these models is challenging due to high computational costs, painfully slow convergence, and overfitting issues.
In this paper, we present \OurMethod{}, a novel approach that enables the efficient and effective training of large models by dividing them into smaller sub-modules which can be trained separately and assembled seamlessly.
A key challenge for implementing this idea is to ensure the compatibility of the independently trained sub-modules.
To address this issue, we first introduce a global, shared meta model, which is leveraged to implicitly link all the modules together, and can be designed as an extremely small network with negligible computational overhead.
Then we propose a module incubation algorithm, which trains each sub-module to replace the corresponding component of the meta model and accomplish a given learning task.
Despite the simplicity, our approach effectively encourages each sub-module to be aware of its role in the target large model, such that the finally-learned sub-modules can collaborate with each other smoothly after being assembled.
Empirically, our method outperforms end-to-end (E2E) training in terms of both final accuracy and training efficiency.
For example, on top of ViT-Huge, it improves the accuracy by \textbf{2.7\%} on ImageNet or achieves similar performance with \textbf{4$\times$} less training time. Notably, the gains are significant for downstream tasks as well (e.g., object detection and image segmentation on COCO and ADE20K).
Code is available at \url{https://github.com/LeapLabTHU/Deep-Incubation}.
\end{abstract}

%% file: intro.tex
\vspace{-.5em}
\section{Introduction}

\begin{figure}[t]\centering
\includegraphics[width=0.9\columnwidth]{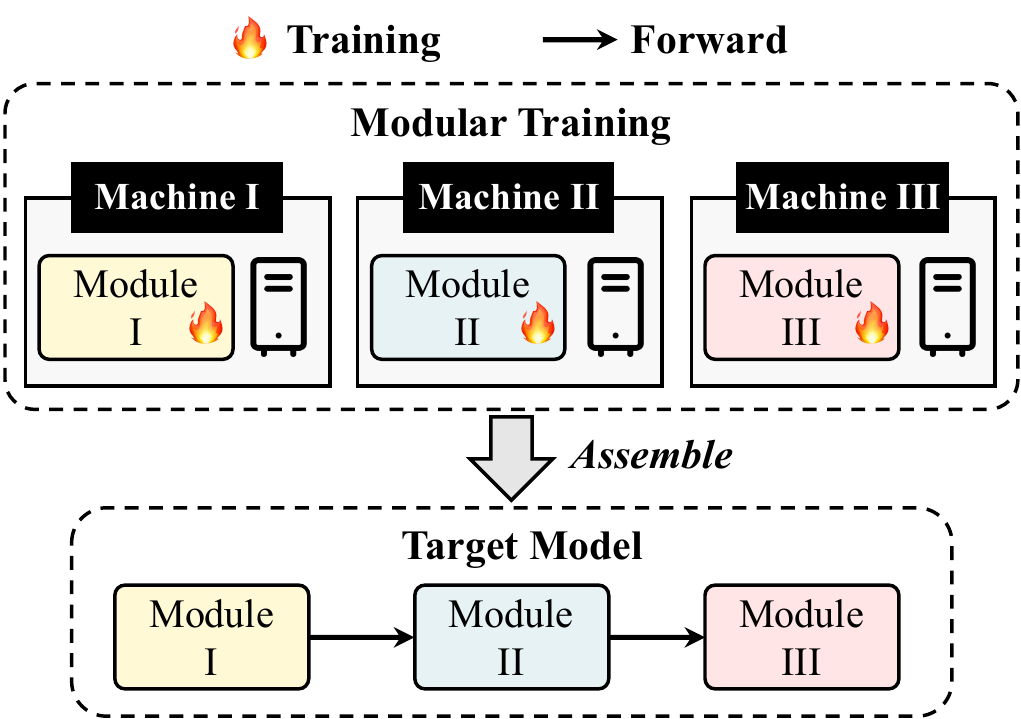}
\vskip -0.12in
\caption{\textbf{An illustration of our idea}.
We first train the sub-modules of a large model fully independently, and then assemble the trained modules to obtain the target model.
}
\label{fig:teaser}
\vskip -0.15in
\end{figure}
Large neural networks have achieved remarkable success across various domains such as natural language understanding~\cite{Radford2019LanguageMA,Brown2020LanguageMA}, computer vision~\cite{chen2020simple,Zhai2021ScalingVT} and reinforcement learning~\cite{Silver2017MasteringCA,berner2019dota}.
In particular, the foundation models~\cite{bommasani2021opportunities,yuan2021florence} heavily rely on large deep learning models to achieve state-of-the-art performance.
However, training large models is challenging in many aspects.
On infrastructure side, centralized resources with strong computational and memory capacities are often required~\cite{kaplan2020scaling,dai2021coatnet,Zhai2021ScalingVT}.
On optimization side, the training process tends to be unstable, difficult to converge, and vulnerable to overfitting~\cite{kaplan2020scaling,dosovitskiy2020image}.
Though recent techniques on GPU memory saving~\cite{rajbhandari2020zero,narayanan2021memory,rajbhandari2021zero} have greatly alleviated the high memory requirements of large models, the optimization issues and the huge training cost still remain challenging.

In this paper, we propose a \emph{divide-and-conquer} strategy to improve the \emph{effectiveness} (better generalization performance) and the \emph{efficiency} (lower training cost) for training large models.
In specific, we divide a large model into smaller sub-modules, train these modules separately, and then assemble them to obtain the final model.
Compared with directly training the whole large network from scratch, starting the learning on top of smaller modules yields a faster and more stable converge process and higher robustness against overfitting.
The independent nature also allows the training of each module to be performed on different machines with no communication needed.
We refer to this paradigm as ``modular training'', and illustrate it in Fig.~\ref{fig:teaser}.

Importantly, designing an effective modular training mechanism is non-trivial, as there exists a dilemma between \textit{independency} and \textit{compatibility}:
although training sub-modules independently enjoys advantages in terms of optimization efficiency and generalization performance, it is challenging to make these modules compatible with each other when assembling them together.
Some preliminary works alleviate this problem by leveraging approximated gradients~\cite{Jaderberg2017DecoupledNI,Czarnecki2017UnderstandingSG,Huo2018DecoupledPB} or local objectives~\cite{belilovsky2019greedy,belilovsky2020decoupled,Wang2021RevisitingLS}, at the price of only achieving partial independency.
However, the modules are still highly entangled during forward propagation, and generally have not exhibited the ability to effectively address the optimization issues faced by training the recently proposed large models (\eg, ViTs, see Tab.~\ref{tab:comp_decoupled}).

In contrast, this paper proposes a \OurMethod{} approach, which not only elegantly addresses this dilemma, but also demonstrates that the training of modern large models can considerably benefit from the \emph{divide-and-conquer} paradigm (see Tab.~\ref{tab:vit_variants} and Fig.~\ref{fig:efficiency}).
Specifically, we first introduce a global, shared meta model, under the goal of implicitly linking all the modules together.
On top of it, we propose a module incubation algorithm that trains each sub-module to replace the corresponding component of the meta model in terms of accomplishing a given learning task (\eg, minimizing the supervised training loss).
This design effectively encourages each sub-module to be aware of its role in the target large model.
As a consequence, even though all the modules are independently trained, we are able to obtain highly compatible sub-modules which collaborate with each other smoothly after being assembled.
Notably, our approach allows deploying an extremely shallow meta model, \eg, only \emph{one} layer per module, with which the computational overhead is negligible, while the performance of the target model will not be affected.
An overview of \OurMethod{} is presented in Fig.~\ref{fig:ours}.

Empirically, extensive experiments of image recognition, object detection and semantic/instance segmentation on competitive benchmarks (\eg, ImageNet-1K~\cite{russakovsky2015imagenet}, ADE20K~\cite{zhou2019semantic} and COCO~\cite{lin2014microsoft}) demonstrate the effectiveness of \OurMethod{}.
For example, with ViT-H, in terms of the generalization performance, \OurMethod{} improves the accuracy by 2.7\% on ImageNet and the mIoU by 3.4 on ADE20K compared to E2E baseline.
From the lens of training efficiency, \OurMethod{} can achieve performance similar to E2E training with 4$\times$ less training cost.

%% file: related.tex
\vspace{-.3em}
\section{Related Work}
\vspace{-.3em}

\paragraph{Decoupled learning} of neural networks is receiving more and more attention due to its biological plausibility and its potential in accelerating the model training process.
Auxiliary variable methods~\cite{taylor2016training,zhang2017convergent,askari2018lifted,li2019lifted} achieve a certain level of decoupling with strong convergence guarantees.
Another line of research~\cite{Bengio2014HowAC,lillicrap2014random,lee2015difference,nokland2016direct} uses biologically motivated methods to achieve decoupled learning.
Using auxiliary networks~~\cite{belilovsky2019greedy,belilovsky2020decoupled,Wang2021RevisitingLS} to achieve local supervision is also a way to achieve decoupling.
However, most above methods focus on decoupling modules during back-propagation, while the modules are still highly entangled during forward propagation.
In contrast, our modular training process completely decouples the modules and optimizes each of them independently.

\paragraph{Model stitching}~\cite{lenc2015understanding,bansal2021revisiting,csiszarik2021similarity} aims to build hybrid models by ``stitching'' model parts from different pre-trained model with stitch layers.
The aim is usually to investigate the internal representation similarity of different neural networks.
A recent work~\cite{yangdeep} also applies model stitching to transfer the knowledge of pre-trained models for downstream tasks.
However, the models obtained by stitching are limited by the architecture and training dataset of the pre-trained models, while our method is a \emph{general training paradigm} that can be applied to any novel architectures and new datasets.

\paragraph{Knowledge distillation}~\cite{Hinton2015DistillingTK,romero2014fitnets,shen2021progressive} trains a small student model to mimic the behavior of a larger model, thus transferring knowledge from the teacher model to the student model and achieves model compression.
This imitative feature has some resemblance to a naïve variant of our method, which is called Module Imitation (see Fig.~\ref{fig:comp} (b)).
However, they are essentially different.
Specifically, the meta models in our work are much smaller than the target models, while in knowledge distillation the teacher networks are typically larger and more powerful than the student networks.
Moreover, our goal is not to compress a large model into a smaller one, but to effectively train a large model with the help of a small meta model.

%% file: method.tex
\vspace{-.3em}

\section{\OurMethod{}}
\begin{figure*}[t]
    \centering
    \includegraphics[width=.8\linewidth]{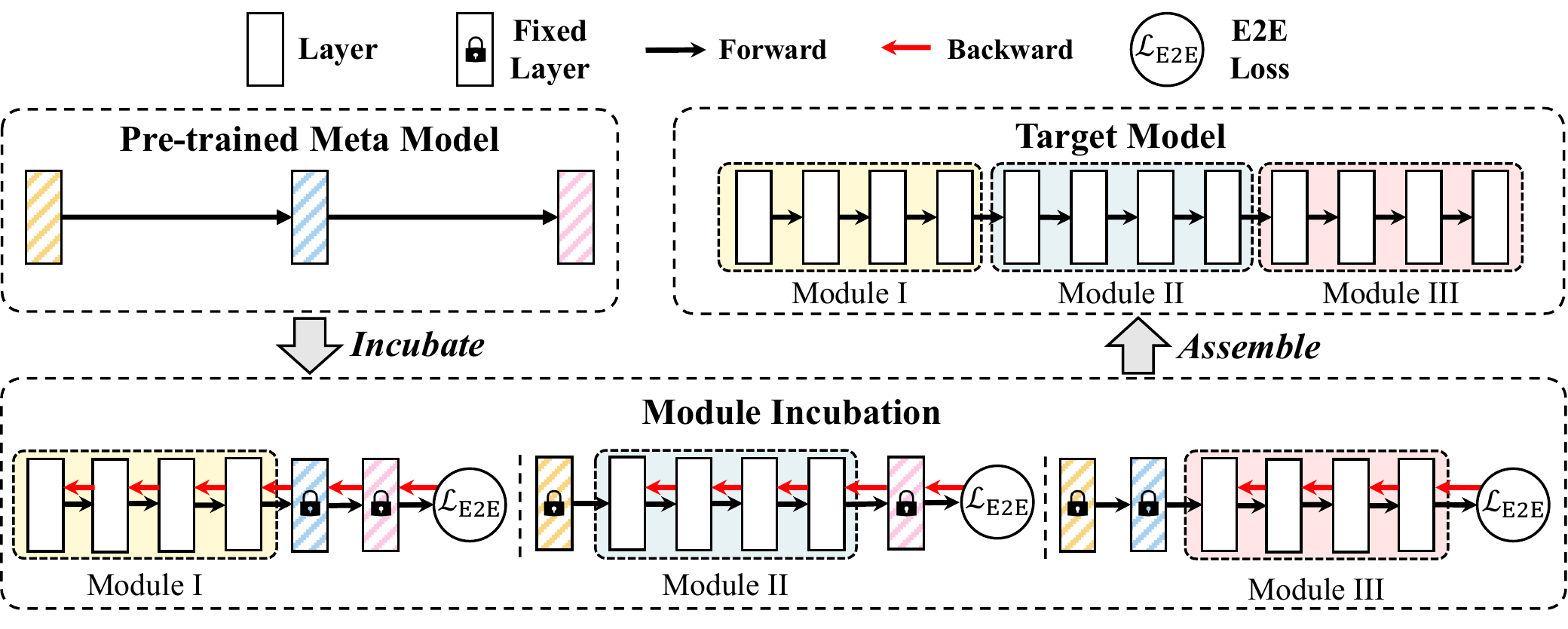}
    \vskip -0.12in
    \caption{
        \textbf{The overall pipeline of \OurMethod{} ($K=3$) .}
        Here, we take a target model with 12 layers as an example, and design a meta model with only one layer per module.
        The meta model is end-to-end pre-trained on the training dataset.
        When training the $i$-th target module (denoted as $M_i$), we simply replace the $i$-th meta layer in the meta model with $M_i$, and train the resulting hybrid network in an end-to-end manner with all meta layers fixed.
        Then, we assemble the trained modules together to obtain the target model.
    }
    \label{fig:ours}
    \vskip -0.12in
\end{figure*}

\vspace{-.3em}
\label{sec:method}
As aforementioned, training large models is typically challenging, \eg, the learning process tends to be unstable, resource/data-hungry, and vulnerable to overfitting.
To tackle these challenges, we propose \OurMethod{}, a divide-and-conquer strategy that improves the \emph{effectiveness} and \emph{efficiency} of large model training.
In this section, we introduce the concept of modular training.
By discussing the difficulties it faces, we present our \OurMethod{} approach and summarize it in Alg.~\ref{algo:module_incub} and Fig.~\ref{fig:ours}.

\paragraph{Modular training} first divides a large model into smaller modules, and then optimizes each module independently.
As modern neural networks are generally constituted by a stack of layers, it is natural to divide the model along the depth dimension.
Formally, given a large target model $\mathcal{M}$ with $n$ layers, we can divide $\mathcal{M}$ into $K(K\le n)$ modules:
\begin{equation}
    \mathcal{M}={M}_K \circ {M}_{K-1} \circ \cdots \circ {M}_1,
\end{equation}
where $\circ$ represents function composition.
Then, each module $M_i$ is trained independently in modular training.

In this way, the cumbersome task of directly training a large model is decomposed into easier sub-tasks of training small modules.
Moreover, these sub-tasks can be distributed to different machines and executed in full parallel, with no communication needed.
After this process, we can simply assemble the trained modules, thus avoiding training the large model directly from scratch.

Therefore, if implemented properly, modular training can be a highly effective and efficient way for large model training.
However, designing a proper modular training mechanism is a non-trivial task.
In the following, we discuss in detail the challenges and present our solutions.

\begin{figure}[t]
    \centering
    \includegraphics[width=\columnwidth]{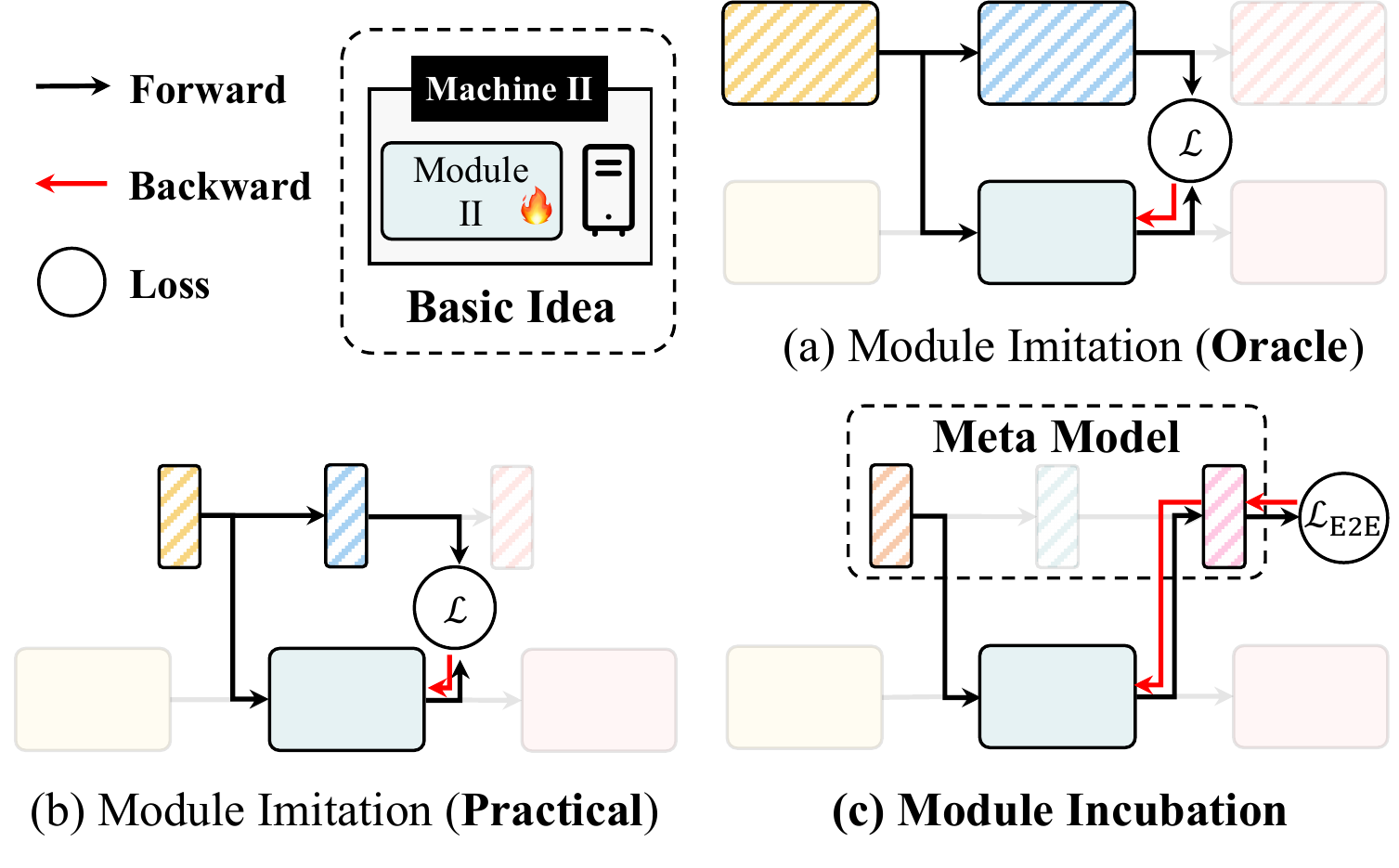}
    \vskip -0.12in
    \caption{\textbf{Comparison of 3 implementations of modular training} when training Module II in the target model ($K=3$).
    In each implementation, the model above is the meta model $\hat{\mathcal{M}}^*$, and the model below is the target model $\mathcal{M}$.
        $\mathcal{L}$ is any measure of distance in feature space, \textit{i.e.}, $L_1$ distance.
        $\mathcal{L}_{\text{E2E}}$ is the original E2E training loss.
        Modules not involved in the training pipeline are greyed out.
    }
    \label{fig:comp}
    \vskip -0.12in
\end{figure}

\paragraph{Dilemma I: independency \vs compatibility.}
At the core of modular training is the requirement of \textit{independency}.
However, if the modules are trained completely unaware of other modules, they may have low \textit{compatibility} between each other, hence negatively affecting the performance of the assembled model.

\paragraph{Solution: meta model.}
We argue the root of the above dilemma is that, the requirement of independency prevents the \textit{explicit} information exchange between modules.
Consequently, the modules cannot adapt to each other during training, causing the incompatible issue.
Driven by this analysis, we propose to address the dilemma by introducing a \textit{global, shared} meta-model $\hat{\mathcal{M}}^*$ to enable \textit{implicit} information exchange between the modules.
Notably, the meta model $\hat{\mathcal{M}}^*$ is designed to have the same number of modules as the target model $\mathcal{M}$:
\begin{equation}
    \hat{\mathcal{M}}^*=\hat{M}^*_K \circ \hat{M}^*_{K-1} \circ \cdots \circ \hat{M}^*_1,
\end{equation}
and is pre-trained on the training dataset.

With the help of the meta model $\hat{\mathcal{M}}^*$, we can easily obtain compatible modules.
For example, we can let each target module $M_i$ imitate the behavior of meta module $\hat{M}^*_i$ by feeding it the same input as $\hat{M}^*_i$, and optimize it to produce feature similar to the output of $\hat{M}^*_i$.
In this way, we can obtain compatible target modules due to the inherent compatibility between the pre-trained meta modules, thus resolving the first dilemma.
We refer to this process of modular training as ``Module Imitation''.
In an oracle case where $\hat{\mathcal{M}}^*$ has the same architecture as $\mathcal{M}$ (Fig.~\ref{fig:comp} (a)), this process can directly produce a good approximate of a well-learned target model when the trained modules are assembled.

\paragraph{Dilemma II: efficiency \vs effectiveness.}
Nevertheless, the solution in Fig.~\ref{fig:comp} (a) may be impractical.
Since our motivation is to train $\mathcal{M}$, it is unreasonable to assume a well-learned meta model $\hat{\mathcal{M}}^*$ of the same size as $\mathcal{M}$ is already available.
More importantly, adopting a large $\hat{\mathcal{M}}^*$ to facilitate modular training can incur unaffordable overhead and make the training process extremely inefficient.
Therefore, in practice, a small meta model needs to be adopted for \textit{efficiency}, as illustrated in Fig.~\ref{fig:comp} (b).
However, small meta models may have insufficient representation learning ability, and thus may limit the performance of the final model.
From this perspective, the meta model should not be too small for the \textit{effectiveness} of modular training.

\paragraph{Solution: module incubation.}
We argue that the above dilemma comes from the inappropriate optimization objective for the target module $M_i$, which is to strictly imitate the meta module $\hat{M}_i^*$.
This objective makes the representation learning ability of $M_i$ bounded by $\hat{M}_i^*$.
Therefore, we propose a ``Module Incubation'' mechanism to better leverage the meta model for modular training.
In specific, instead of letting $M_i$ strictly imitate $\hat{M}_{i}^*$, we encourage ${M}_i$ to cooperate with the meta model $\hat{\mathcal{M}}^*$ to attain a task-oriented learning goal.
Formally, we replace the $i$-th module in the meta model $\hat{\mathcal{M}}^*$ with $M_i$, obtaining a hybrid network $\tilde {\mathcal{M}} ^{(i)}$:
\begin{equation}
    \label{eq:local_train}
    \tilde {\mathcal{M}} ^{(i)} = \hat M_K^* \circ \cdots \circ \hat M_{i+1}^* \circ \textcolor{red}{M_i} \circ \hat M_{i-1}^* \circ \cdots \circ \hat M_1^*.
\end{equation}
Then we fix $\hat M_j^*(j\neq i)$, and thus the outputs of $\tilde {\mathcal{M}} ^{(i)}$ corresponding to the input $\bm{x}$ can be defined as a function of $M_i$, namely:
\begin{equation}
    \bm{x} \to \tilde {\mathcal{M}} ^{(i)}(\bm{x}; M_i).
\end{equation}
Finally, we can directly minimize an end-to-end loss $\mathcal{L}_{\textnormal{E2E}}(\cdot)$ with respect to $\tilde {\mathcal{M}} ^{(i)}(\bm{x}; M_i)$:
\begin{equation}
    \label{eq:our_method}
    \mathop{\textnormal{minimize}}\limits_{M_i} \ \ \  \mathcal{L}_{\textnormal{E2E}}\left(y,\ \   \tilde {\mathcal{M}} ^{(i)}(\bm{x}; M_i)\right),
\end{equation}
where $y$ is the label of the input $\bm{x}$. Here, $\mathcal{L}_{\textnormal{E2E}}(\cdot)$ can be defined conditioned on the task of interest.
In this paper, we mainly consider the standard cross-entropy loss in the context of classification problems.
The above process can be seen as using the pre-trained meta model $\hat{\mathcal{M}}^*$ to ``incubate'' the target module $M_i$, and thus we call this way of modular training ``Module Incubation''.

Unlike Module Imitation, here we enforce $M_i$ to cooperate with $\hat{M}_j^*(j\neq i)$ to accomplish the final task.
Therefore, $M_i$ is encouraged to take full advantage of its potential.
Since $M_i$ is often larger than $\hat M_{i}^*$, it can acquire stronger ability than $\hat M_{i}^*$ in terms of representation learning.
In contrast, the ability of $M_i$ is generally limited by the insufficient meta module $\hat M_{i}^*$ in Module Imitation.
Empirical evidence is also provided in Fig.~\ref{fig:LT_ours_KD} to support this point.

Interestingly, we find that smaller meta models actually bring \textit{better} performance in Module Incubation (see Fig.~\ref{fig:abl_meta_depth}).
This intriguing phenomenon provides a favorable solution to the second dilemma, \textit{i.e.}, we can directly use the shallowest meta model to incubate the modules.
In our implementation, to get both efficiency and effectiveness, we simply design the meta model to have only \textit{one} layer\footnote{Following~\cite{dosovitskiy2020image}, we use `layer' in a general sense to represent the basic building block of a model , \eg, a transformer encoder layer in ViT.} per module.

\begin{algorithm}[t]
    \small
    \caption{The \OurMethod{} Algorithm}
    \label{algo:module_incub}
    \begin{algorithmic}[1]
        \Require
        Initialize the target model $\mathcal{M}=M_K \circ M_{K-1} \circ \cdots \circ M_1$; Training dataset $\mathcal{D}$
        \State Initialize a meta model $\hat {\mathcal{M}}$ with $K$ modules.
        \State Pre-train $\hat {\mathcal{M}}$ on $\mathcal{D}$ to obtain $\hat {\mathcal{M}}^*$
        \For{$i=1$ {\bfseries to} $K$}
            \Comment{Can be executed in parallel}
            \State Construct $ \tilde {\mathcal{M}}^{(i)}$ by replacing $\hat M_i ^*$ in $\hat {\mathcal{M}} ^*$ with $M_i$
            \State Minimize $\mathcal{L}_{\textnormal{E2E}} \left(y,\ \   \tilde {\mathcal{M}} ^{(i)}(\bm{x}; M_i)\right)$ on $\mathcal{D}$ to obtain $M_i^*$
        \EndFor
        \State Assemble the target model $\mathcal{M}^{\text{assm}} = {M}_K^{*} \circ {M}_{K-1}^* \circ \cdots \circ {M}_1^*$
        \State Fine-tune $\mathcal{M}^{\text{assm}}$ on $\mathcal{D}$ to obtain the final model $\mathcal{M}^{*}$
    \end{algorithmic}
\end{algorithm}

\begin{figure*}[t!]
  \begin{center}
    \begin{minipage}{0.58\linewidth}
        \tablestyle{1.6mm}{1.15}
        \begin{tabular}{cccc|ccc|cc}
            \multirow{2}{*}{model} & \multirow{2}{*}{\shortstack{image \\size}} & \multirow{2}{*}{FLOPs} & \multirow{2}{*}{\#param} & \multirow{2}{*}{\shortstack{E2E-ViT\\\cite{dosovitskiy2020image}}} & \multirow{2}{*}{\shortstack{E2E\\\cite{he2022masked}}} & \multirow{2}{*}{\shortstack{E2E-DeiT\\\cite{Touvron2021TrainingDI}}} & \multirow{2}{*}{\shortstack{DeiT \\+ ours}} & \multirow{2}{*}{$\Delta$} \\
            &&&&&&&\\\shline
            \multirow{2}{*}{\shortstack{ViT-B}} & $224^2$ & 17.6G & \multirow{2}{*}{\shortstack{87M}} & -  & 82.3 & 81.8& \textbf{82.4} & \textbf{\textcolor{up}{+0.6}}\\
            & $384^2$ & 55.5G &  & 77.9  & - & 83.1& \textbf{84.2} & \textbf{\textcolor{up}{+1.1}}\\\hline
            \multirow{2}{*}{\shortstack{ViT-L}} & $224^2$ & 61.6G & \multirow{2}{*}{\shortstack{304M}} & -  & 82.6 & 81.4$^\dagger$& \textbf{83.9}& \textbf{\textcolor{up}{+2.5}}\\
            & $384^2$ & 191.1G &  & 76.5  & - & 83.3$^\dagger$& \textbf{85.3} & \textbf{\textcolor{up}{+2.0}}\\\hline
            \multirow{2}{*}{\shortstack{ViT-H}} & $224^2$ & 167.4G & \multirow{2}{*}{\shortstack{632M}} & -  & 83.1 & 81.6$^\dagger$& \textbf{84.3} & \textbf{\textcolor{up}{+2.7}}\\
            & $392^2$ & 545.3G & & -  & - & 83.4$^\dagger$& \textbf{85.6} & \textbf{\textcolor{up}{+2.2}}
        \end{tabular}
        \vspace{-.9em}
            \captionof{table}{
        \textbf{Training large ViT models on ImageNet-1K}.
            For wall time training efficiency comparison, see Fig.~\ref{fig:efficiency}.
            $\dagger$: Our reproduced baseline. \label{tab:vit_variants}
        }
    \end{minipage}
    \hspace{0.05ex}
    \begin{minipage}{0.4\linewidth}
        \vspace{1.2em}
        \tablestyle{1.3mm}{1.15}
        \begin{tabular}{cc|ccc|c}
            dataset & model & \textcolor{gray}{E2E}                                & DGL~\cite{belilovsky2020decoupled}                            & InfoPro~\cite{Wang2021RevisitingLS}                               & ours \\\shline
            \multirow{3}{*}{C100} & ResNet-110  & \textcolor{gray}{71.1}                                & 69.2                            & 71.2                               & \textbf{73.0} \\
            & DeiT-T-32 & \textcolor{gray}{72.8}                                & 72.0                            & 73.3                               & \textbf{75.3} \\
            & DeiT-T-128 & \textcolor{gray}{69.4}                                & 70.9                            & 73.2                               & \textbf{77.2} \\\hline
            IN-1K & ViT-B & \textcolor{gray}{81.8} & -& 81.0  & \textbf{82.4} \\
        \end{tabular}
        \vspace{-1em}
                \captionof{table}{
            \textbf{Comparison with decoupled learning methods.} The results on InfoPro~\cite{Wang2021RevisitingLS} and DGL~\cite{belilovsky2020decoupled} are based on our implementation.
            C100: CIFAR-100, IN-1K: ImageNet-1K.
        }
        \label{tab:comp_decoupled}
    \end{minipage}
  \end{center}\vskip -0.3in
\end{figure*}

\begin{figure*}[!t]\centering
\includegraphics[width=\linewidth]{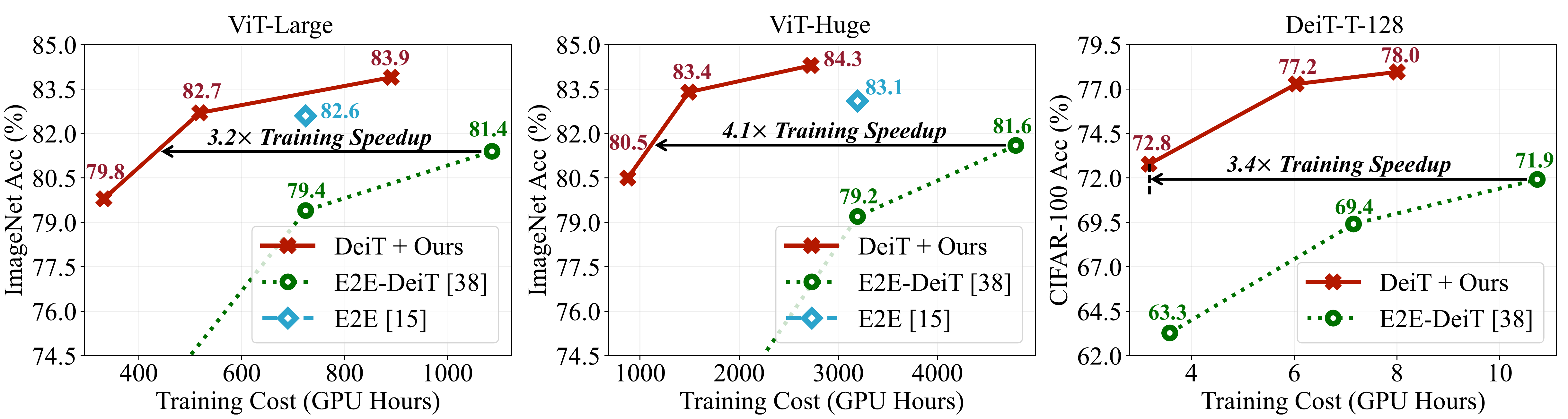}
\vskip -0.12in
\caption{\textbf{Training efficiency (accuracy \vs training wall-time)} of ViT-L (\textit{left}), ViT-H (\textit{middle}) on ImageNet-1K and DeiT-T-128 on CIFAR-100 (\textit{right}). Different points correspond to different training budgets (\textit{i.e.}, with varying numbers of epochs).
        The training cost is measured in A100 GPU Hours. We also present detailed convergence curves of ViT-Huge in Fig.~\ref{fig:train_curve}.
}
\label{fig:efficiency}
\vskip -0.12in
\end{figure*}
\paragraph{Assemble the target model.}
After all the modules $M_i (i \in \{1, \dots , K\})$ are trained, we obtain the target model by assembling them:
\begin{equation}
    \mathcal{M}^{\text{assm}} = {M}_K^{*} \circ {M}_{K-1}^* \circ \cdots \circ {M}_1^*,
\end{equation}
where $M_i^{*}$ denotes the modular-trained target module. Then we fine-tune $\mathcal{M}^{\text{assm}}$ to obtain the final model $\mathcal{M}^*$.
Importantly, this fine-tuning process does \emph{not} downplay the importance of modular training.
To demonstrate this issue, we can consider E2E training as a special case of \OurMethod{} where the proportion of fine-tuning is 100\%.
However, such 100\% fine-tuning significantly degrades the test accuracy  (see: Tab.~\ref{tab:vit_variants}). 
Only when the modular training stage is introduced, a dramatically improved generalization performance can be achieved (see: Fig.~\ref{fig:abl_stage1_ratio}).
This demonstrates that the major gain of our method comes from the modular training algorithm rather than the fine-tuning process.

The overall pipeline of \OurMethod{} is summarized in Alg.~\ref{algo:module_incub} and Fig.~\ref{fig:ours}.




%% file: experiments.tex
\section{Experiments}
\label{sec:experiments}

This section presents a comprehensive experimental evaluation on ImageNet-1K \cite{russakovsky2015imagenet}, COCO~\cite{lin2014microsoft}, ADE20K~\cite{zhou2019semantic} and CIFAR \cite{Krizhevsky2009LearningML} to validate the effectiveness of \OurMethod{}.

\paragraph{Setups.}
We adopt the widely used training recipe of DeiT~\cite{Touvron2021TrainingDI} as our default training configurations, with small modifications: for large models like ViT-L and ViT-H, we adapt the stochastic depth rate accordingly following~\cite{touvron2021going} by setting it to 0.5 (ViT-L) and 0.6 (ViT-H) for both our fine-tuning phase and E2E baselines, and extend the warmup epochs to 20 following~\cite{liu2022convnet}.
The E2E baselines trained with default configuration serve as our main baseline, which is denoted as E2E-DeiT.
For the simplicity of our method, we intentionally adopt the \textit{same} hyper-parameters for both our modular training and the fine-tuning phase as E2E-DeiT, except that we disable warmup in the fine-tuning phase.
Therefore, we refer to our method as DeiT + Ours.

We keep $K\!=\!4$ for modular training and evenly divide the target models.
The depth of meta models is set to 4, which is the shallowest possible meta model.
We simply perform modular training for 100 epochs and fine-tuning for 100 epochs.
This configuration is selected for an optimal overall performance.
Notably, shorter fine-tuning still produces favorable results (see Fig.~\ref{fig:sensitivity}).
We pre-train meta models for 300 epochs with the same configurations as E2E-DeiT.
Note that the pre-training cost of meta models is cheap compared to the overall training cost due to the shallowness of meta models (see Fig.~\ref{fig:train_curve}).
The schedules of the E2E baselines are the same as their original papers.

\subsection{Main Results}

\paragraph{Training large models on ImageNet-1K.}
Since the results of ViT-L and ViT-H are not reported in DeiT \cite{touvron2021going}, and directly adopting the original training configurations results in optimization issues (\textit{i.e.} NaN loss), we report our reproduced baselines.
Besides the re-adjusted stochastic depth rates, we also adopt the LAMB~\cite{You2020LargeBO} optimizer and a uniform stochastic depth rate following~\cite{touvron2021going} to further improve E2E training.

As shown in Tab.~\ref{tab:vit_variants}, our method consistently improves model performance on the top of E2E-DeiT for all three ViT variants.
The advantage is more pronounced for larger models.
On ViT-H, our method outperforms E2E-DeiT by \textbf{2.7\%}.
The advantage continues when the models are fine-tuned at higher resolution, where our method outperforms E2E-DeiT by 2.0\% and 2.2\% for ViT-L and ViT-H, respectively.
We also compare \OurMethod{} with the recently proposed improved E2E baselines in \cite{he2022masked}, where a systematically hyper-parameter search is performed on training configurations.
Notably, this comparison places our method at a \textit{disadvantage} since we directly adopt the configurations of E2E-DeiT, which may be sub-optimal for our method.
Even so, \OurMethod{} still performs better.

\paragraph{Comparison with decoupled training methods.}
We compare our method with two strong decoupled training methods: InfoPro~\cite{Wang2021RevisitingLS} and DGL~\cite{belilovsky2020decoupled} with both ViTs and CNNs.
We adopt two DeiT-Tiny~\cite{Touvron2021TrainingDI} models with a depth of 32 and 128 and a ResNet-110 ($K=3$) on CIFAR and ViT-B on ImageNet-1K.
For models on CIFAR, we train our method for 200 (modular training) + 100 (fine-tuning) and other baselines for 400 epochs.
The results are presented in Tab.~\ref{tab:comp_decoupled}.
Our method consistently outperforms the other state-of-the-art decoupled training methods.

\paragraph{Higher computational efficiency for training.}
\begin{figure}[!h]\centering
\includegraphics[width=\linewidth]{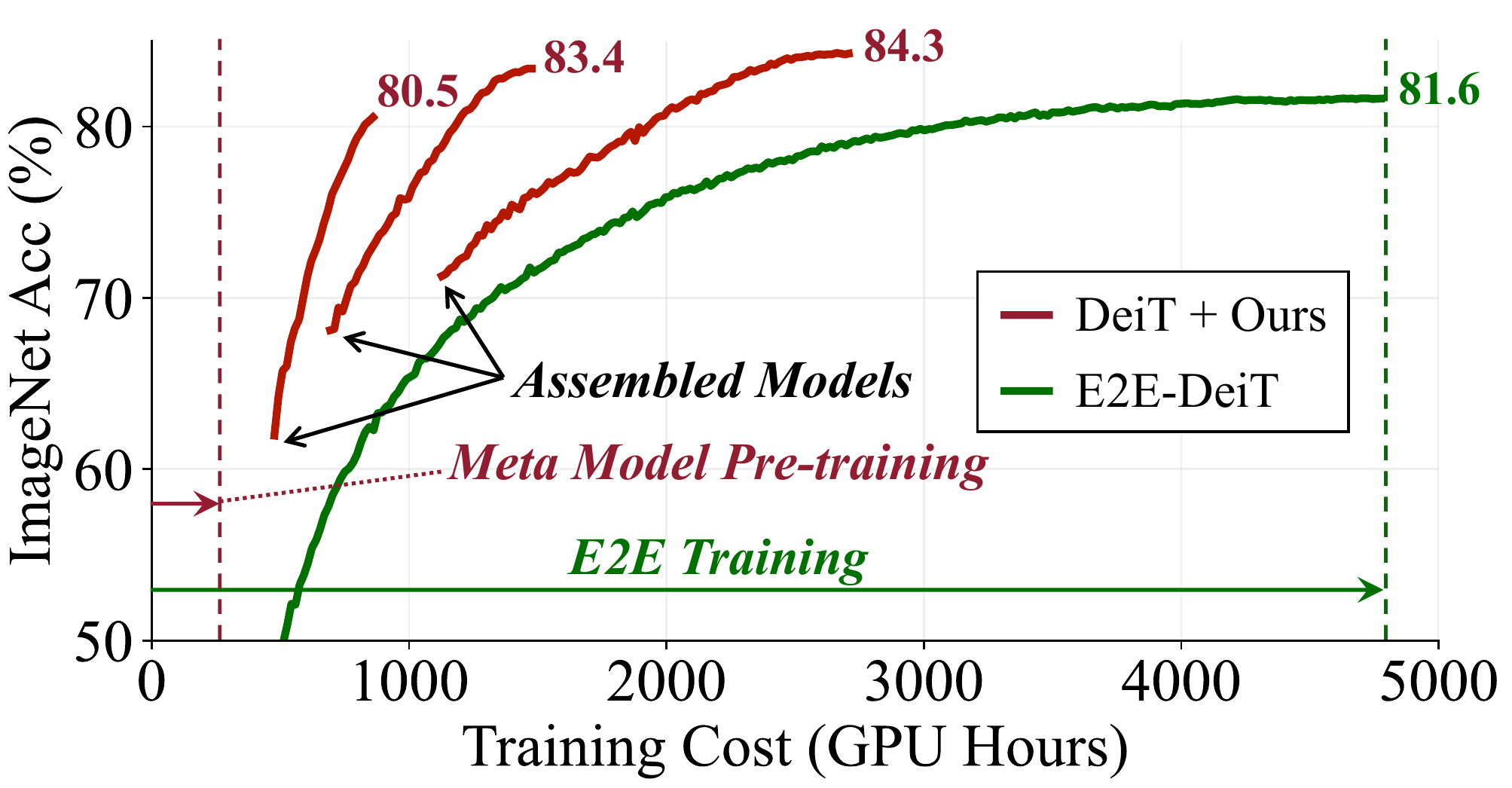}
\vskip -0.15in
\caption{\textbf{Training curves of ViT-H.}
Our method (with different training budgets) is compared with E2E-DeiT~\cite{Touvron2021TrainingDI}.
}
\label{fig:train_curve}
\vskip -0.15in
\end{figure}
In Fig.~\ref{fig:efficiency}, we present a more comprehensive comparison of the training efficiency between our method and the E2E baselines.
We adjust the training cost budget by varying the number of epochs.
One can observe that our method shows a better efficiency-accuracy trade-off than all E2E baselines, including the recently proposed improved E2E baselines~\cite{he2022masked}.

For fair comparisons, we further discuss the benefits of our method on training efficiency by comparing it with E2E-DeiT since they adopt the same training configurations.
On ViT-L and ViT-H, our method requires 3.2$\times$ and 4.1$\times$ less training cost, respectively, while achieving similar performance compared to E2E-DeiT.
We also present detailed convergence curves of ViT-H in Fig.~\ref{fig:train_curve}.
For our method, we plot the convergence curve starting from the assembled models.
Notably, the starting points of our convergence curves are \textit{higher} than the convergence curve of E2E training.
This demonstrates the high compatibility between the modules trained by our method.

\paragraph{Downstream tasks.}
To further demonstrate the effectiveness of our method, we evaluate our ImageNet-1K pre-trained models on 2 common downstream tasks: COCO object detection \& instance segmentation and ADE20K semantic segmentation.
COCO~\cite{lin2014microsoft} object detection and instance segmentation dataset has 118K training images and 5K validation images.
We employ our pre-trained models on the commonly used Mask R-CNN~\cite{he2017mask} framework with 1$\times$ and 3$\times$ training schedule.
ADE20K~\cite{zhou2019semantic} is a popular dataset for semantic segmentation with 20K training images and 2K validation images.
We employ our pre-trained backbone on the widely used segmentation model UperNet~\cite{xiao2018unified} and train it for 80k steps.
For ViT-Huge, we interpolate the patch embedding filters from 14$\times$14$\times$3 to 16$\times$16$\times$3 to fit the input image sizes of downstream tasks.
As shown in Tab.~\ref{tab:coco} and \ref{tab:ade20k}, our pre-trained backbones achieve consistent improvement over E2E trained counterparts with a \textit{lower} pre-training cost.

\begin{table}[!t]
    \begin{raggedright}
        \tablestyle{.9mm}{1.25}
        \begin{tabular}{c|llllll}
            & \multicolumn{6}{c}{Mask R-CNN, 1$\times$ schedule} \\
            & AP$^b$  & AP$^b_\text{50}$ & AP$^b_\text{75}$ & AP$^m$ & AP$^m_\text{50}$ & AP$^m_\text{75}$ \\\shline
            E2E & 44.1 & 67.7 & 48.5 & 40.4 & 64.2 & 43.1 \\
            Ours & \dtplus{45.7}{1.6} & \dtplus{69.6}{1.9} & \dtplus{50.3}{1.8} & \dtplus{41.8}{1.4} & \dtplus{66.1}{1.9} & \dtplus{44.7}{1.6} \\
        \end{tabular} \\
        \vspace{.5em}
        \begin{tabular}{c|llllll}
            & \multicolumn{6}{c}{Mask R-CNN, 3$\times$ schedule} \\
            & AP$^b$  & AP$^b_\text{50}$ & AP$^b_\text{75}$ & AP$^m$ & AP$^m_\text{50}$ & AP$^m_\text{75}$ \\\shline
            E2E & 47.0 & 69.3 & 51.3 & 42.1 & 66.2 & 45.2 \\
            Ours & \dtplus{48.6}{1.6} & \dtplus{71.2}{1.9} & \dtplus{52.9}{1.6} & \dtplus{43.8}{1.7} & \dtplus{68.1}{1.9} & \dtplus{47.2}{2.0} \\
        \end{tabular} \\
    \end{raggedright}
    \vspace{-.7em}
    \caption{\textbf{Object detection and instance segmentation on COCO val2017}.
    Here, we adopt ViT-L as backbone to compare our \OurMethod{} pre-training with E2E~\cite{Touvron2021TrainingDI} and use Mask R-CNN~\cite{he2017mask} as detector.
    For the pre-training cost of ViT-L, see Tab.~\ref{tab:ade20k}.
    \label{tab:coco}
    }
    \vskip -0.1in
\end{table}

\begin{table}[!t]
    \tablestyle{1.6mm}{1.25}
    \begin{tabular}{ccc|lll}
        \multirow{2}{*}{model} & \multirow{2}{*}{method} & \multirow{2}{*}{\shortstack{pt. cost\\ (GPU hours)}} & \multicolumn{3}{c}{UperNet, 80k training steps}   \\
        &&& mAcc. & mIoU  & mIoU$^\dagger$ \\\shline
        \multirow{2}{*}{ViT-L} & E2E & 1.09K   & 58.5 & 47.0 & 47.8 \\
        & Ours   & 0.89K &\dtplus{60.8}{2.3} & \dtplus{49.2}{2.2} & \dtplus{50.0}{2.2} \\\hline
        \multirow{2}{*}{ViT-H}  & E2E & 4.79K    & 58.2 & 46.5 &  47.2     \\
        & Ours & 2.72K  & \dtplus{61.0}{2.8} & \dtplus{49.9}{3.4} & \dtplus{50.6}{3.4}   \\
    \end{tabular}
    \vspace{-.7em}
    \caption{\textbf{Semantic segmentation on ADE20K}.
    Here, we test our pre-trained models compared to the E2E trained ones~\cite{Touvron2021TrainingDI} with UperNet~\cite{xiao2018unified}.
        $^{\dagger}$ denotes the multi-scale test setting with flip augmentation.
    }
    \label{tab:ade20k}
    \vskip -0.1in
\end{table}
\paragraph{Higher data efficiency.}
\begin{table}[!h]
    \tablestyle{.8mm}{1.25}
    \begin{tabular}{c|cc|cc}
    {training} & \multicolumn{2}{c|}{top-1 acc.} & \multicolumn{2}{c}{training loss} \\
    data & E2E-DeiT~\cite{Touvron2021TrainingDI} & DeiT + Ours & E2E-DeiT~\cite{Touvron2021TrainingDI} & DeiT + Ours \\\shline
    \!100\% IN-1K\! & 81.8 & \textbf{82.4 \textcolor{Highlight}{(+0.6)}}  & 2.63 & 2.69 \\
    \!50\% IN-1K\! & 74.7 & \textbf{78.6 \textcolor{Highlight}{(+3.9)}} & 2.34& 2.55  \\
    \!25\% IN-1K\! & 65.7 & \textbf{72.9 \textcolor{Highlight}{(+7.2)}}   & 2.09& 2.41 \\
    \end{tabular}
    \vspace{-.7em}
    \caption{\textbf{Training ViT-B with fewer training samples} (IN-1K: ImageNet-1K).
    Here, we sample 2 class-balanced subsets from ImageNet-1K.
    The training loss in the last epoch is also reported.
    }
    \label{tab:data_efficiency}
    \vskip -0.1in
\end{table}
Another important advantage of our method is its higher data efficiency, \textit{i.e.}, it is able to dramatically outperform the E2E baselines when training data are relatively scarce.
To demonstrate this, we sample two class-balanced subsets of ImageNet-1K, containing 25\% and 50\% of the training data, and train ViT-B on them.
The training cost is kept the same as full-set training by using the same number of training iterations.
Besides Top-1 accuracy, we also report the training loss in the last epoch.

The results are reported in Tab.~\ref{tab:data_efficiency}.
It can be seen that the gain of our method is more pronounced in lower data regimes, since our method is less prone to overfitting.
As data become scarce, we can observe that the \textit{training loss} of E2E training quickly decreases, while the \textit{validation accuracy} also drops, showing a clear trend of overfitting.
In contrast, our method counters this trend with a much slower decrease in training loss, and achieves significantly higher validation accuracy than E2E training.
For example, when only 25\% of ImageNet-1K is available, it outperforms the DeiT baseline by \textbf{7.2\%}.

\subsection{Designing Deeper Models}
With our proposed method, we can further explore an interesting question: \textit{in current transformer models, is the ratio of depth \vs width optimal?}
The answer may be true in the context of E2E learning.
Previous work~\cite{gong2021vision,zhou2021deepvit} conjecture that deeper ViTs trained in an end-to-end manner may have the over-smoothing problem, hindering their performance and hence it is not suggested to make ViTs deeper than their current design.

\begin{figure}[!h]\centering
\vspace{-.7em}
\includegraphics[width=0.7\linewidth]{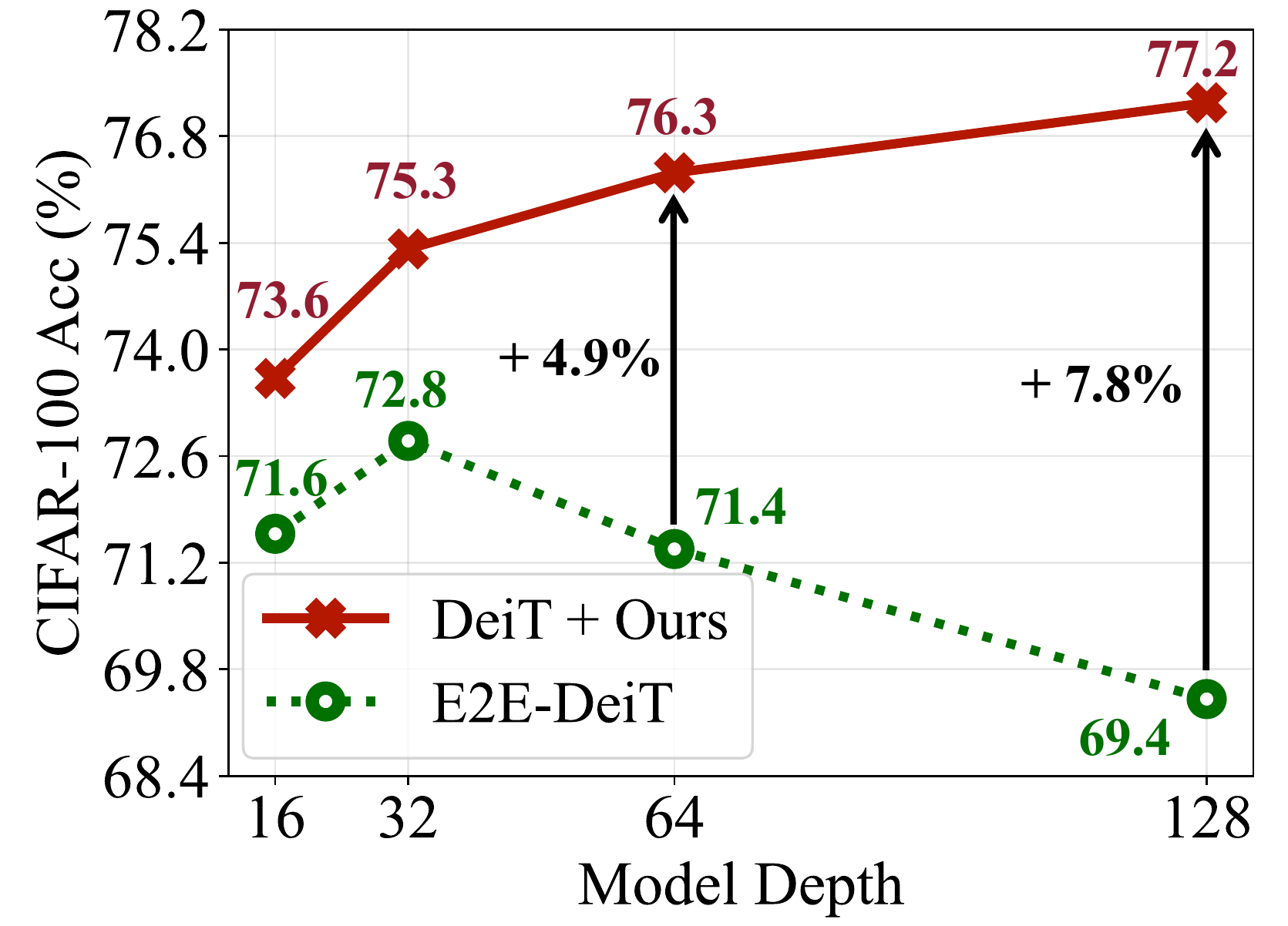}
\vskip -0.12in
\caption{\textbf{Increasing depth} of DeiT-Tiny.
Our method is able to train deeper ViTs without optimization issues.
}
\label{fig:tiny_inc_depth}
\vskip -0.1in
\end{figure}

To investigate this problem, we progressively increase the depth of a DeiT-Tiny in Fig.~\ref{fig:tiny_inc_depth}, and train them on CIFAR-100 to evaluate their performance.
One can observe that the performance of E2E learning quickly saturates when depth increases to 32 and then starts to decrease as the model depth further increases.
However, the same phenomenon does not occur with our method. The models trained by our method show no sign of saturation in performance and outperform E2E counterparts by increasingly larger margins as the model gets deeper. In other words, our method is able to train deeper ViTs more effectively.

\begin{table}[h]
    \vspace{-.3em}
    \tablestyle{.9mm}{1.15}
    \begin{tabular}{lcccc|cc}
        \multirow{2}{*}{model} & \multirow{2}{*}{\!FLOPs\!} & \multirow{2}{*}{\!\#param\!} & \multirow{2}{*}{\!depth\!} & \multirow{2}{*}{\!width} & \multicolumn{2}{c}{top-1 acc.} \\
        &&&&&  E2E-DeiT~\cite{Touvron2021TrainingDI} & DeiT + Ours \\\shline
        ViT-B & 17.6G & 87M & 12 & 768 & 81.8 & \textbf{82.4} \\
        ViT-B-DN & 17.7G & 85M & 24 & 540 & 78.7 & \textbf{82.7} \\
    \end{tabular}
    \vspace{-.7em}
    \caption{\textbf{Training deep-narrow models.}
    Here, a deep-narrow version of ViT-B is designed (denoted as `ViT-B-DN') by doubling the depth of ViT-B with the FLOPs unchanged.
    }
    \label{tab:deep_narrow}
    \vskip -0.15in
\end{table}
Intrigued by this observation, we conjecture that our proposed method may allow the designing of more efficient ViTs by further increasing the model depth.
Therefore, we create a deep-narrow version of ViT-Base model (denoted ViT-B-DN) by doubling the depth and adjusting the width accordingly to keep the inference cost (\textit{i.e.}, FLOPs) unchanged.
As shown in Tab.~\ref{tab:deep_narrow}, the deep-narrow version of ViT-B performs significantly worse than its original configuration when trained in an E2E manner.
However, when trained by our method, the deep-narrow version actually outperforms the original one, giving an additional 0.3\% improvement in the final performance.
This provides an alternative solution for scaling up transformer models.

\subsection{Discussion}
In this section, we provide a more comprehensive analysis of our method.
Unless mentioned otherwise, we use DeiT-T-128 as our target model and conduct experiments on CIFAR-100 dataset.
\label{sec:discussion}

\begin{figure}[!t]\centering
\vskip -0.1in
\includegraphics[width=0.7\linewidth]{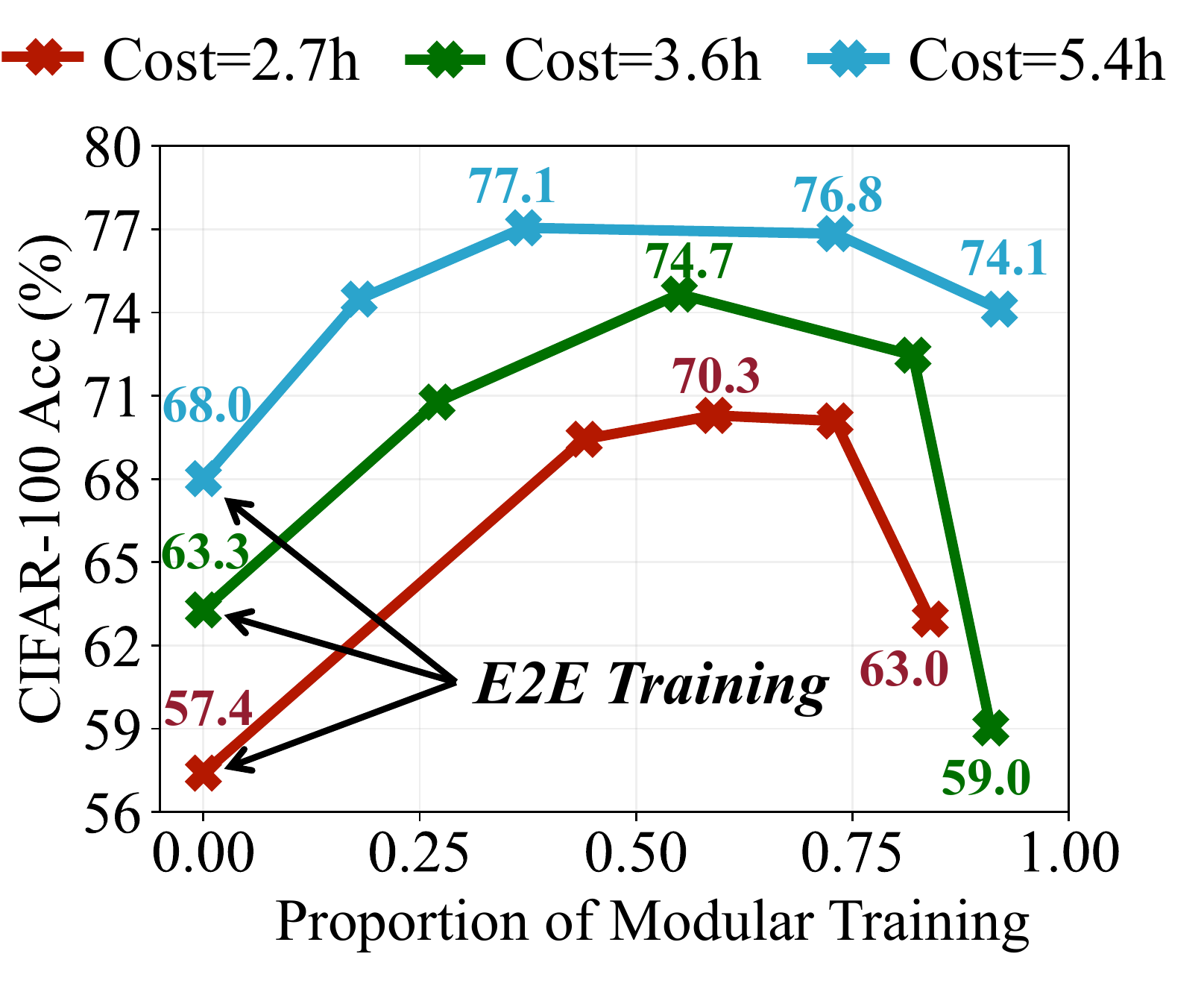}
\vskip -0.15in
\caption{\textbf{Proportion of modular training}.
The proportion is measured by the wall-clock time of modular training in the whole pipeline of \OurMethod{}.
When the proportion of modular training is zero, our method reduces to E2E training.
}
\label{fig:abl_stage1_ratio}
\vskip -0.15in
\end{figure}
\paragraph{Ablation on modular training.}
We first try directly replacing modular training in our method with E2E training, and keep the fine-tuning stage unchanged (denoted as `E2E training + tuning').
This results in a staged E2E training process that adopts a cosine annealing schedule with restart.
The results are shown below:
\begin{center}
    \tablestyle{7pt}{1}
    \begin{tabular}{ccc}
        modular training + tuning (ours) & E2E training + tuning & E2E \\\shline
        \textbf{77.2} & 67.9 & 69.4 \\
    \end{tabular}
\end{center}
This indicates that the gain of our method does not come from the staged training process itself, which even underperforms the E2E baseline.

We further study the importance of modular training by varying its proportion in the whole process, while keeping the overall training cost unchanged.
As shown in Fig.~\ref{fig:abl_stage1_ratio}, starting from E2E training (the proportion of modular training is zero), the overall performance considerably improves as more computation is allocated to modular training.
Furthermore, \OurMethod{} outperforms E2E training within a wide range of the proportions, which also demonstrates its robustness.

\paragraph{Ablation on meta model.}
In our module incubation formulation, we pre-train and fix the meta model for incubating modules.
The table below shows the effect of the pre-training and the fixing operation on the final performance:
\begin{center}
    \vskip -0.07in
    \tablestyle{5pt}{1}
    \begin{tabular}{ccc}
        pre-trained, fixed (ours) & pre-trained, tunable & random init., tunable \\\shline
        \textbf{77.2} & 76.4 & 70.9 \\
    \end{tabular}
\end{center}
Thus, pre-training and fixing are both beneficial to the overall performance, with pre-training being more important.
This is reasonable since the meta model is to facilitate the compatibility between independently trained modules, and thus needs to be: 1) pre-trained to ensure its own layers' compatibility and 2) fixed to be consistent when incubating different modules.
Note the gain from meta model pre-training does \emph{not} comes from the pre-trained meta model itself, which only has an accuracy of 64.9\%.

\begin{figure}[!h]\centering
\includegraphics[width=.7\columnwidth]{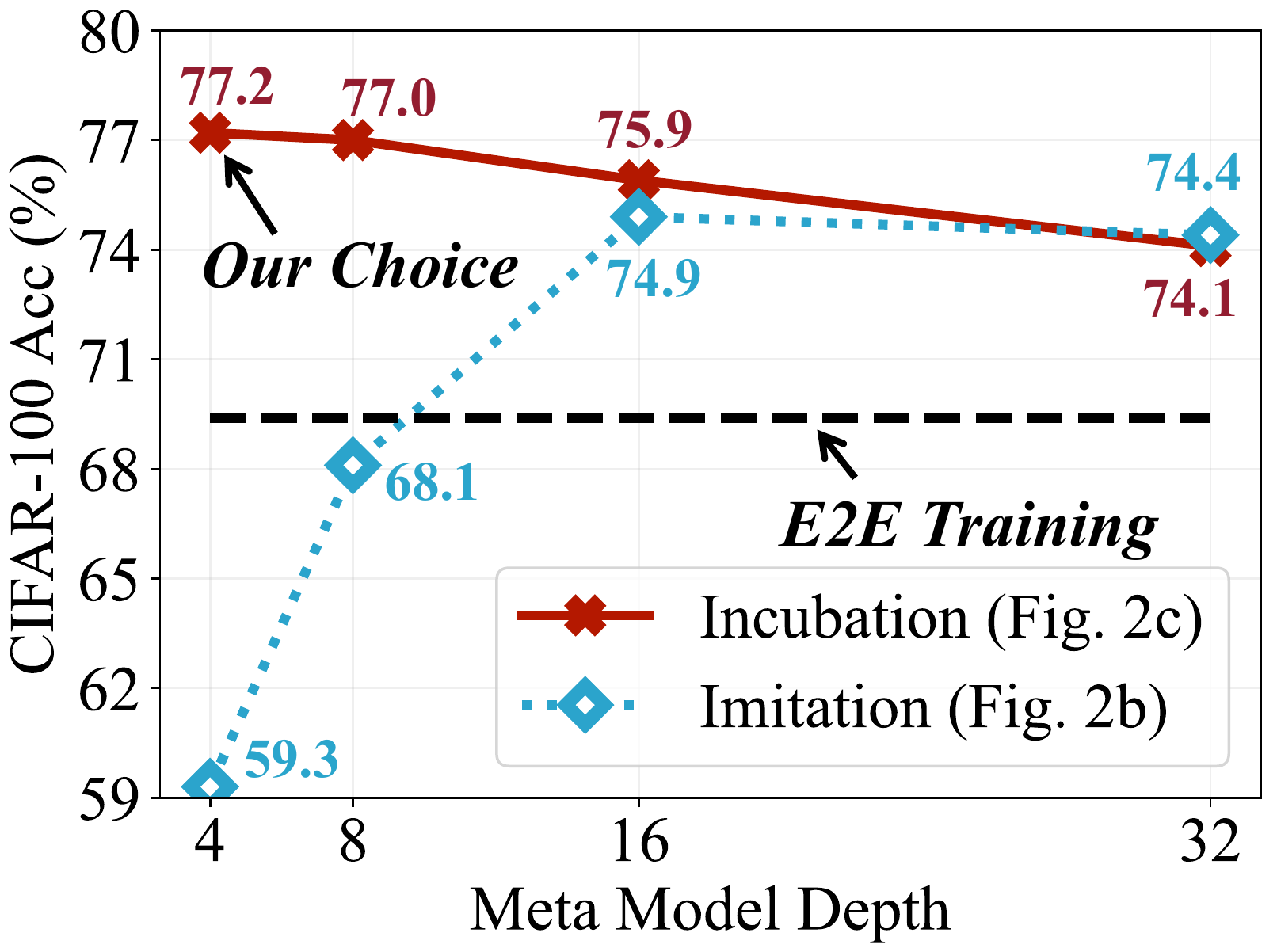}
\vskip -0.12in
\caption{\textbf{Depth of the meta model}.
We perform modular training with meta models of varying depths.
Two ways of implementation, \ie, Module Imitation (Fig.~\hyperref[fig:comp]{2b}) and Module Incubation (ours, Fig.~\hyperref[fig:comp]{2c}), are compared.
}
\label{fig:abl_meta_depth}
\vskip -0.2in
\end{figure}

We further study the effect of meta model depth in Fig.~\ref{fig:abl_meta_depth}.
The accuracy of our method is depicted in a red line, where the horizontal axis denotes the number of layers of the meta model
An intriguing observation can be obtained, \textit{i.e.}, our method achieves high accuracy even with a surprisingly shallow meta model (\eg, 4 layers, one for each module).
One possible explanation for this phenomenon is that, during the module incubation process, adopting shallower meta models makes the supervision information flow more easily toward the target module, and thus the target modules can be trained more thoroughly and converge faster.

\paragraph{Comparison with module imitation.}
Fig.~\ref{fig:abl_meta_depth} also presents the results of Module Imitation (Fig.~\ref{fig:comp} (b)), where we adopt $L_1$ distance as the loss function in feature space.
It can be seen that our method consistently outperforms Module Imitation, especially when the meta model is small.
This is aligned with our intuition in Sec.~\ref{sec:method} that the cooperative nature of Module Incubation prevents the representation learning power of $M_i$ from being limited by an insufficient meta model.
\begin{figure}[!h]\centering
\vspace{-.3em}
\includegraphics[width=.7\columnwidth]{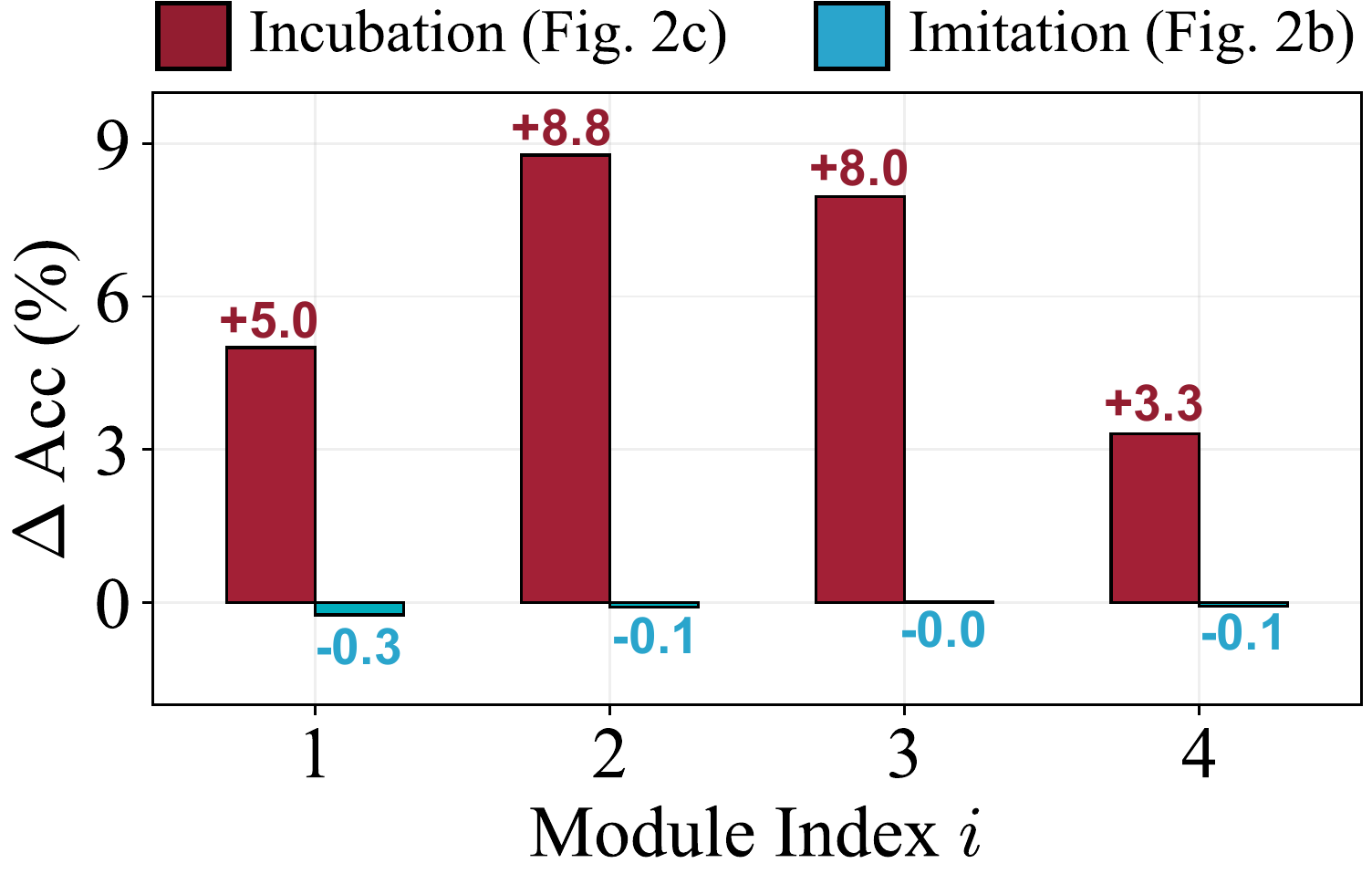}
\vskip -0.12in
\caption{\textbf{Accuracy gain} when replacing a meta module $\hat{M}_i^*$ in the meta model with target module $M_i^*$ trained by different methods.
}
\label{fig:LT_ours_KD}
\vskip -0.1in
\end{figure}

We also explicitly measure how well a trained target module $M_i^*$ supports a meta model to learn representations by replacing the meta module $\hat{M}_i^*$ in the meta model with $M_i^*$.
The accuracy gain of this hybrid model over the original meta model, which is DeiT-T-4, is evaluated.
As the results in Fig.~\ref{fig:LT_ours_KD} show, the modules trained by Module Incubation (ours) do provide better support for the meta model by leveraging its stronger ability in representation learning.

\begin{figure}[!t]
    \vspace{-.7em}
    \hspace{10ex}
    \includegraphics[width=.7\columnwidth]{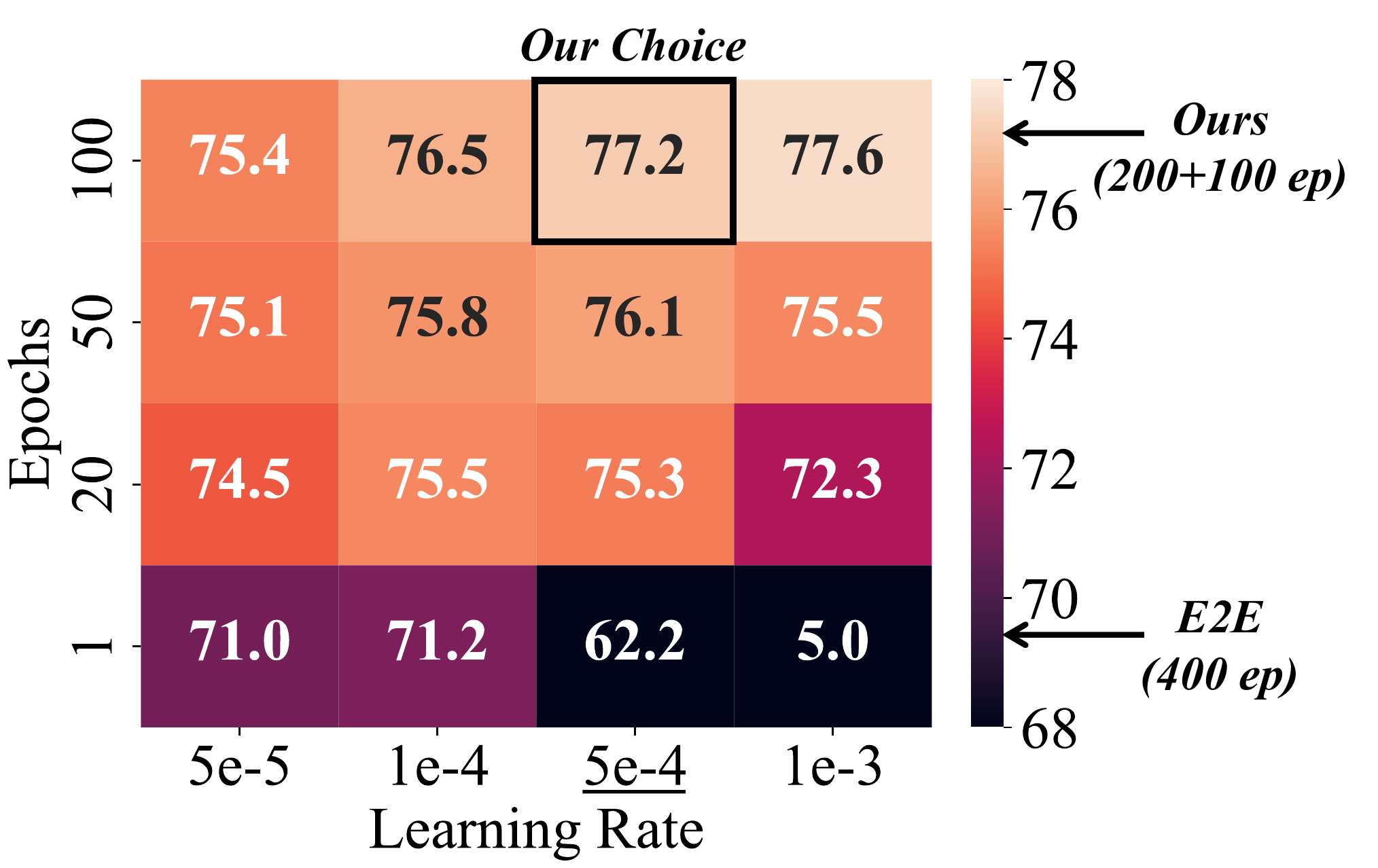}
    \vskip -0.12in
    \caption{\textbf{Sensitivity test} on the hyper-parameters of fine-tuning on CIFAR-100.
    The default learning rate is \underline{underlined}.
    }
    \label{fig:sensitivity}
    \vskip -0.2in
\end{figure}

\paragraph{Sensitivity test.}
We further conduct a sensitivity test on the hyper-parameters for fine-tuning the assembled model, namely, the epochs and the learning rate for fine-tuning.
The results are shown in Fig.~\ref{fig:sensitivity}, where we use DeiT-T-128 as the target model.
Three important observations can be obtained.
\textit{First}, our method can outperform E2E training even if the model is only fine-tuned for \textbf{one} epoch (71.2\% for ours \vs 69.4\% for E2E), which clearly demonstrates the necessity of our modular training process.
\textit{Second}, the majority of the performance gain can be obtained by fine-tuning the assembled model for a short period (\eg, 20 epochs), and further prolonging the fine-tuning phase gives diminishing returns.
\textit{Third}, the performance of our method is generally robust to the choice of the learning rate of fine-tuning.
For a moderate period of fine-tuning, directly choosing the default learning rate is enough.
Therefore, for all the experiments, we do not tune this learning rate to keep the simplicity of our method.

\paragraph{Number of modules $K$.}
Finally, we also present our study on $K$, which is the number of modules when we divide a target model.
The results are presented below:
\begin{center}
    \vskip -0.1in
    \tablestyle{2.4mm}{1.1}
    \begin{tabular}{l|cccc|c}
        model  & $K=2$  & $K=4$  & $K=8$  & $K=16$ & E2E  \\\shline
        DeiT-T-32 &  72.3 & \textbf{76.1} & 75.6 &  75.6    & 72.8 \\
        DeiT-T-256 & 70.9 & 76.7 & \textbf{77.2} & 75.0   & 66.9
    \end{tabular}
    \vskip -1in
\end{center}
It can be seen that the optimal value of $K$ differs for models of different depths, and the deeper model prefers a larger $K$.
This is reasonable since gradient vanishing and other optimization problems get more severe for deeper models, and thus a finer division of the model is needed.

%% file: conclusion.tex
\section{Conclusion}
\label{sec:conclusion}
This paper presented \OurMethod{}, which trains a large model in a divide-and-conquer manner.
We leveraged a shared, lightweight meta model to implicitly link all modules together.
By ``incubating'' the modules with the meta model, we effectively encouraged each module to be aware of its role in the target large model, and thus the trained modules can collaborate with each other smoothly after they are assembled.
Extensive experiments demonstrated that \OurMethod{} can outperform E2E training dramatically in terms of generalization performance, training efficiency and data efficiency.

%% file: acknowledgement.tex
\section*{Acknowledgement}
This work is supported in part by the National Key R\&D Program of China under Grant 2019YFC1408703, the National Natural Science Foundation of China under Grant 62022048, THU-Bosch JCML and Beijing Academy of Artificial Intelligence.